\documentclass{article}

\usepackage{microtype}
\usepackage{graphicx}
\usepackage{subfigure}
\usepackage{booktabs} % for professional tables
\usepackage{multirow}
\usepackage{amsmath}
\usepackage{color}
\usepackage{amssymb}

\usepackage{hyperref}

\usepackage[accepted]{icml2019}

\icmltitlerunning{Predicting Treatment Initiation from Clinical Time Series Data via Graph-Augmented Time-Sensitive Model}

\begin{document}

\twocolumn[
\icmltitle{Predicting Treatment Initiation from Clinical Time Series Data \\ via Graph-Augmented Time-Sensitive Model}

% It is OKAY to include author information, even for blind
% submissions: the style file will automatically remove it for you
% unless you've provided the [accepted] option to the icml2019
% package.

% List of affiliations: The first argument should be a (short)
% identifier you will use later to specify author affiliations
% Academic affiliations should list Department, University, City, Region, Country
% Industry affiliations should list Company, City, Region, Country

% You can specify symbols, otherwise they are numbered in order.
% Ideally, you should not use this facility. Affiliations will be numbered
% in order of appearance and this is the preferred way.
\icmlsetsymbol{equal}{*}

\begin{icmlauthorlist}
\icmlauthor{Fan Zhang}{iqvia}
\icmlauthor{Tong Wu}{umn}
\icmlauthor{Yunlong Wang}{iqvia}
\icmlauthor{Yong Cai}{iqvia}
\icmlauthor{Cao Xiao}{iqvia}
\icmlauthor{Emily Zhao}{iqvia}
\icmlauthor{Lucas Glass}{iqvia}
\icmlauthor{Jimeng Sun}{gatech}

\end{icmlauthorlist}

\icmlaffiliation{iqvia}{IQVIA Inc., Plymouth Meeting, PA, USA}
\icmlaffiliation{umn}{Department of Biomedical Engineering, University of Minnesota Twin Cities, Minneapolis, MN, USA}
\icmlaffiliation{gatech}{Georgia Institute of Technology, Atlanta, GA, USA}
%\icmlaffiliation{ed}{School of Computation, University of Edenborrow, Edenborrow, United Kingdom}

\icmlcorrespondingauthor{Yunlong Wang}{Yunlong.Wang@iqvia.com}

% You may provide any keywords that you
% find helpful for describing your paper; these are used to populate
% the "keywords" metadata in the PDF but will not be shown in the document
\icmlkeywords{Machine Learning, ICML}

\vskip 0.3in
]

% this must go after the closing bracket ] following \twocolumn[ ...

% This command actually creates the footnote in the first column
% listing the affiliations and the copyright notice.
% The command takes one argument, which is text to display at the start of the footnote.
% The \icmlEqualContribution command is standard text for equal contribution.
% Remove it (just {}) if you do not need this facility.

\printAffiliationsAndNotice{}  % leave blank if no need to mention equal contribution
%\printAffiliationsAndNotice{\icmlEqualContribution} % otherwise use the standard text.

\begin{abstract}
Many computational models were proposed to extract temporal patterns from clinical time series for each patient and among patient group for predictive healthcare. However, the common relations among patients (e.g., share the same doctor) were rarely considered. In this paper, we represent patients and clinicians relations by bipartite graphs addressing for example from whom a patient get a diagnosis. We then solve for the top eigenvectors of the graph Laplacian, and include the eigenvectors as latent representations of the similarity between patient-clinician pairs into a time-sensitive prediction model. We conducted experiments using real-world data to predict the initiation of first-line treatment for Chronic Lymphocytic Leukemia (CLL) patients. Results show that relational similarity can improve prediction over multiple baselines, for example a 5\% incremental over long-short term memory baseline in terms of area under precision-recall curve.  

% A core issue with time series is investigating their similarity, which is central to important tasks such as forecasting, mining, and clustering.
% In this paper, we use graph analysis to extract latent relational patient similarity to improve prediction. 
% In particular, 

%\js{add some concrete performance numbers}

% and show that the similarity features can be used to improve the performance of time series based prediction.
% As a real-world example, we investigate the prediction of initiating first-line treatment for Chronic Lymphocytic Leukemia (CLL) patients from their clinical time series data.

% Experimental results show that the proposed treatments can lead to more accurate prediction for CLL patients.

\end{abstract}

\section{Introduction}

% \cx{for the introduction, I'll suggest start with talking about massive health data are collected, which provide great opportunity for health analytics [then cite a few]. Among the health data, clinical time series is one major type. Then instead of saying "forecasting, or prediction, is arguably the most sought-after one", we mention over the years, many models were proposed to make disease detection, progression modeling, and patient subtyping based on clinical time series [then cite a few].}
Recent years there has been an explosion in the amount of digital information growth in electronic health records, which provides great opportunities for applications such as health analytics and clinical informatics \cite{xiao2018opportunities,topol2019}.
Among the electronic health data, clinical time series is one major type.
Over the years, many computational models have been proposed for disease detection \cite{choi2016using,Choi2018-cj}, disease progression \cite{bai2018interpretable}, and patient subtyping \cite{baytas2017patient,che2017rnn} based on clinical time series data.
% A core issue regarding studies of time series is determining the similarity of one time series with respect to another, which is central to tasks such as forecasting, mining, and clustering.
% Over the years several similarity measures for time series data have been proposed, such as cross-correlation, dynamic time wrapping \cite{sakoe1990dynamic}, and edit distance \cite{fischer2013fast,chen2005robust}.
% The proposed similarity measures are expected to be able to handle high-dimensional time series data while being fast, scalable, and efficient \cite{serra2014empirical}.

% \cx{the second paragraph: we first define disease progression modeling task (in general sense). after defining this task, we talk about what characteristics of clinical ts that can be leverage in solving this task. and then cite existing works to briefly mention what has been done. }
A disease progression model is designed to predict the development of potential treatments for many slowly progressing diseases, e.g. Alzheimer's disease, by detecting more granular stages as compared to those defined in clinical diagnosis \cite{sukkar2012disease}.
Longitudinal clinical time series along with related patient or physician data is important in informing disease progression patterns, and motivate many deep learning based disease progression models including recurrent neural network (RNN), attention model, and graph embedding \cite{choi2016doctor,choi2016retain,bai2018interpretable,che2017rnn,suresh2017clinical}.

% patients' demographics, diagnoses, prescriptions, and clinical notes, most of which are time-stamped and organized in longitudinal structures.
% Recent works that leveraged clinical time series data for predicting disease progression are mainly based on deep learning models for example recurrent neural network (RNN), attention model, and graph embedding. \cite{xiao2018opportunities,choi2016doctor,choi2016retain,bai2018interpretable,che2017rnn,suresh2017clinical}.

%In recent years, the fast development of deep learning has also allowed modeling disease progression via deep neural networks that can better handle the massive clinical datasets with hierarchical structures, and achieve improved results \cite{shickel2018deep,yadav2018mining}.

% \cx{the third paragraph: we highlight what is missing from existing work that we propose to address with this work. this paragraph provides intuition to motivate our method.} 
%\cx{in this paper, lets define "relational similarity or relational relevancy" instead of just calling it "similarity" since "patient similarity has been long studied for personalized prediction"}
Despite these initial success, the relational structure between patients and clinicians has been overlooked and is of great potentials to enhance prediction accuracy. 
%We are interested in utilizing the relational structures between patients with the same disease and their visited clinicians.
The assumption is that patients with the same disease who visit the same clinicians tend to receive similar treatments, which can be arguably attributed to that clinicians follow a set of common medical knowledge, and more likely make similar decisions for patients with the same disease. 

\begin{figure}[t]
\begin{center}
\centerline{\includegraphics[width=\columnwidth]{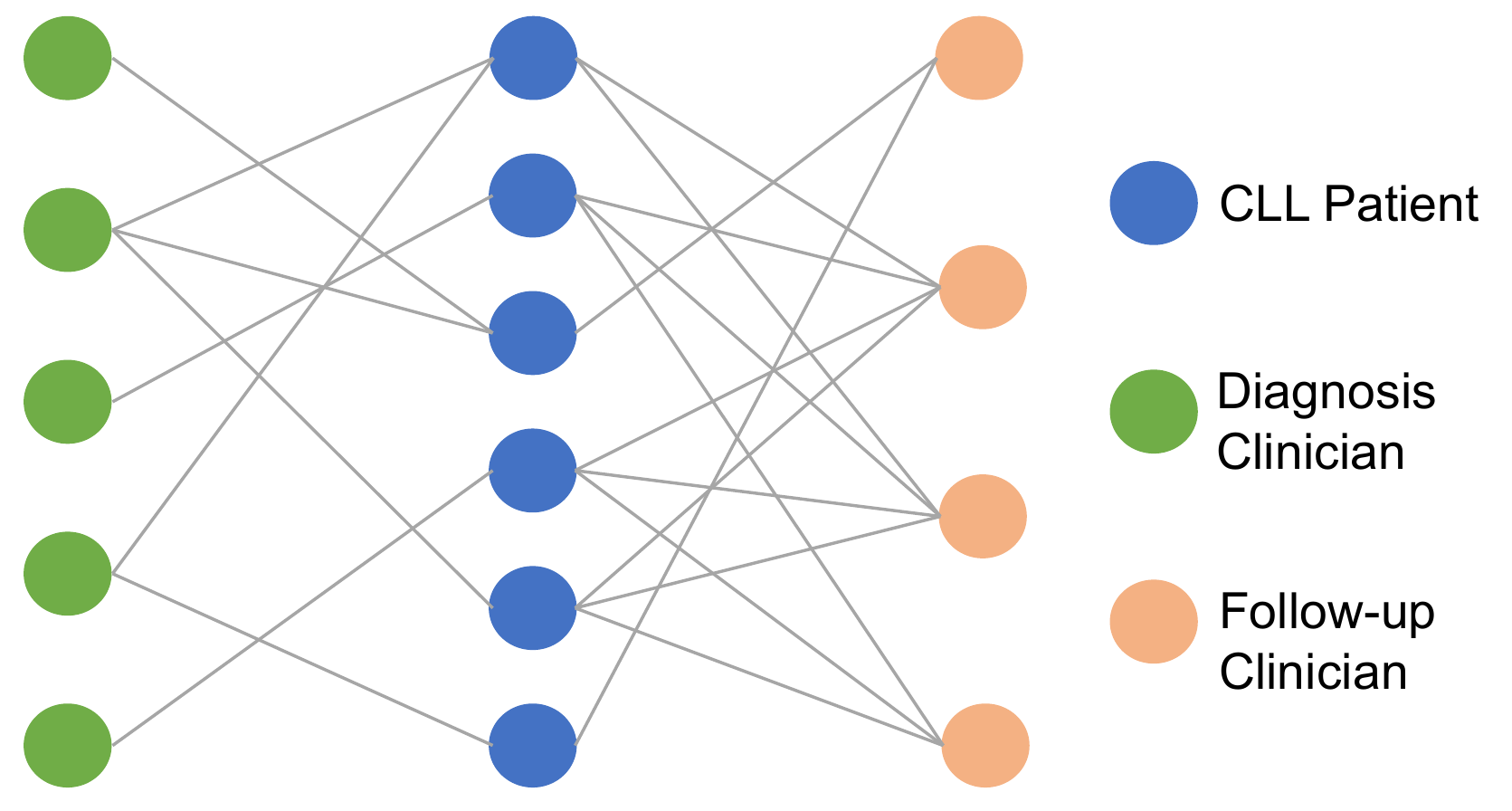}}
\caption{Illustration of the bipartite graph between CLL patients and diagnosis/follow-up clinicians. Follow-up clinicians are restricted to oncologists and hematologists. Diagnosis and follow-up clinicians can overlap.}
\label{bg}
\end{center}
\vspace{-20pt}
\end{figure}

We propose to model the patient-clinician relational structure as a bipartite graph, in which the two disjoint sets of vertices represent the patients and the clinicians, respectively; the set of edges records the number of visits made by each patient with each clinician.
% \cx{add the motivation for creating two separate graphs}
We note that many patients visit different clinicians for diagnosis and follow-up treatments.
Thus we create two graphs, with one for patients and their diagnosis clinicians, and another for patients and their follow-up clinicians.
% \cx{replace the current figure with a figure that has two previously mentioned graphs on it, and indicate their relations in the model/task in the figure }
An example of the graphs using patients' clinical data from IQVIA's prescription and claim database is given in Figure \ref{bg}.
Here we show six patients diagnosed with Chronic Lymphocytic Leukemia (CLL), and their visited clinicians for both diagnosis and follow-up treatments.
% a type of cancer in which the bone marrow makes too many lymphocytes.
%For CLL patients, there is normally an observational period before starting first-line treatment.
%We restrict the follow-up clinicians to oncologists and hematologists for better relevance.
We then apply spectral graph analysis to solve for the top eigenvectors of the graph Laplacian, and take the eigenvectors as latent representations of the similarity between patient-clinician pairs.
The extracted features capture the latent proximity of patients who visit the same clinicians, and can enhance the prediction accuracy of a variety of disease progression models.

% \cx{i have revised the first sentence. the second sentence/contribution also needs revision. the concepts "local sparse features" was never defined before. so either define it in previous paragraphs or be be specific here.}.
As a key contribution, to our best knowledge, this work is the first disease progression/detection model that leverages graph theory to exploit the relational similarity of clinician-visiting from patients' clinical time series data.
Using the patients' clinical medical data from IQVIA's database, we show that the proposed similarity features can improve the performance of a wide choice of machine learning models, from XGBoost \cite{chen2016xgboost} to more recent deep learning based models, e.g. convolutional neural network (CNN) and long short-term memory (LSTM).

% A primary goal of oncology pharmaceutical market research is to develop a quantitative prediction model that leverages machine learning methods and patients’ medical records to perform automated and accurate prediction of disease progression, thus can help raise awareness of treatment options to the relevant clinicians, and also beneficial to early disease intervention.

% To accurately and quickly identify patient before the treatment decisions even being made would empower foresight ability and competitive edge for marketers, allowing the marketers to focus on the right physicians with the right patients at the right time, before the treatment decisions are made \cite{a2015metaboloepigenetic,cravens2006strategic}.

%\vspace{-10pt}
\section{Related Works}

% \cx{in fact, the related works should only focus on treatment prediction. no need to talk about disease progression modeling. lets also limit the discussin within deep learning models.}
Medical treatment prediction is a core research task of disease progression modeling.
Recently, many deep learning models have made rapid advancements on this topic.
In \cite{choi2016retain}, a two-level attention model was designed to detect influential past visits and significant clinical variables for better prediction accuracy and interpretability.
In \cite{choi2017gram}, a graph-based attention model was proposed to extract hierarchical information from medical oncologies and improve RNN-based rare disease prediction.
In \cite{ma2017dipole}, a bi-directional RNN was designed to remember information of both the past and future visits based on three attention mechanism to measure the relationship of different visits for prediction.
In \cite{che2017rnn}, a RNN architecture through dynamically matching temporal patterns was proposed to learn the similarity between two longitudinal patient record sequences for personalized prediction of Parkinson's Disease. 
Other similar approaches have also been proposed \cite{ma2018kame,ma2018general}.

% However, longitudinal medical records often exhibit long-term, complex dependencies and non-linear dynamics; they also suffer from being noisy, sparse, and irregular.
% Given the complexities and imperfections of the clinical data, the performance of regression model or HMM are suboptimal due to their oversimplified assumptions.
%For instance, in HMM disease progression is treated as a Markov process in which the medical treatment in each visit depends only on the treatments conducted in the previous visit, whereas in reality, the diagnosis of a certain disease may rely on many previous examinations.
% Many deep learning based models, such as recurrent neural networks and long short-term memory with enhanced long-term memorization, have been widely applied to model disease progression in recent years and made improvements on prediction accuracy \cite{choi2016doctor,choi2016using,bai2018interpretable,che2017rnn}.
% Instead of more complicated network structures, we focus on designing a new input feature that measures the similarity between time series in a low-dimensional latent space, and can improve the prediction accuracy of arbitrary machine learning models.

%\vspace{-5pt}
\section{Methods}

\begin{table}[t]
\caption{List of notations defined in this paper.}
\centering
\label{notation}
\begin{tabular}{p{3cm} p{4.4cm}}
\toprule[.8pt]
Notation & Description \\\toprule[.5pt]
$\mathcal{D}$ & Clinical medical records \\
$\mathbf{S}$ & Patients' sequences of visits \\
$\mathcal{I}$ & Patients' demographics \\
$\mathcal{U}, \mathcal{V}$ & Sets of clinicians and patients. \\
$M=\mathcal{|U|}, N=\mathcal{|V|}$ & Sizes of $\mathcal{U}$ and $\mathcal{V}$. \\
$\mathcal{E}=\{\{w_{ij}\}_{i=1}^{M}\}_{j=1}^{N}$ & Set of edges connecting $\mathcal{U}$ and $\mathcal{V}$; $w_{ij}$ is the count of patient $j$ visiting clinician $i$. \\
$\mathcal{G=\{U, V, E\}}$ & Bipartite graph for patients and their visited clinicians \\
$\mathbf{A}$ & Adjacency matrix of $\mathcal{G}$ \\
$\mathbf{D}$ & Degree matrix of $\mathbf{A}$ \\
$\mathbf{L}$ & Laplacian matrix of $\mathcal{G}$ \\
$\{\mathbf{X}_i\}_{i=1}^K$ & Relational similarity features \\
\bottomrule[.8pt]
\end{tabular}
\vspace{-10pt}
\end{table}

\subsection{Problem Formulation}
% \cx{add this section, here add a notation table; define data and task mathematically}
%In this work we are interested on predicting initiation of first-line treatment for CLL patients.
We let $\mathcal{D}$ denote patients' clinical medical records.
For a patient $p$, $\mathcal{D}^{(p)}=\{\mathbf{S}^{(p)}, \mathcal{I}^{(p)}\}$, where $\mathbf{S}^{(p)}=[\mathbf{s}_1^{(p)},\mathbf{s}_2^{(p)},\dots,\mathbf{s}_T^{(p)}]$ is the sequence of visits and $\mathcal{I}^{(p)}$ includes demographics and related medical features.
Each visit $\mathbf{s}_t^{(p)}$ consists of information such as diagnoses, procedures, prescriptions, visited clinicians, etc.
The goal of prediction is to learn $f:\mathcal{D}^{(p)} \mapsto y^{(p)}$, where $y^{(p)}$ is the label indicating if the patient will start treatment in the next time window.

We model the patient-clinician relation as a bipartite graph $\mathcal{G=\{U,V,E\}}$ over $\mathcal{D}$, where $\mathcal{U}$, $\mathcal{V}$ are the sets of clinicians and patients, $\mathcal{E}$ is the set of edges connecting $\mathcal{U}$ and $\mathcal{V}$, with each weight $w_{ij}$ denoting the count of patient $j$ visiting clinician $i$.
Table \ref{notation} lists the notations we used in the paper.

% We extract data from IQVIA database including hundreds of millions longitudinal prescription (Rx) and medical claims (Dx). 
% In this study, we focus on Chronic Lymphocytic Leukemia (CLL), for which there is a ``watch-and-wait'' period before patients start first-line treatment. 
% IQVIA receives 2 billion prescription claims per year with history from January 2004 with coverage up to 88\% for the retail channel, 50-70\% for traditional and specialty mail order, and 40\% for long-term care. 
% This information represents activities during the prescription transaction including product, provider, payer, and geography. 
% In this study, we pulled relevant Rx/Dx data from January 2010 to January 2018.

\subsection{Extracting Relational Similarities from Patients' Clinical Time Series via Graph Laplacian}\label{bigraph}

% \cx{need one sentence that connects the motivation and techniques. Then start to describe the following step-by-step practical guide.}
The algorithm for extracting relational similarity is motivated by spectral clustering \cite{ng2002spectral}, in which data points are considered as nodes of a similarity graph and mapped to a low-dimensional space where they can be segregated to form clusters.

Firstly, we construct $K$ bipartite graphs $\{\mathcal{G}_i\}_{i=1}^K$ from $\mathcal{D}$, where each $\mathcal{G}_i=\{\mathcal{U}_i, \mathcal{V}, \mathcal{E}_i\}$ and $\mathcal{E}_i=\{\{w_{ij}\}_{i=1}^{|\mathcal{U}_i|}\}_{j=1}^{|\mathcal{V}|}$.
We let $M=|\mathcal{U}_i|$ and $N=|\mathcal{V}|$ for simplicity.
$\mathcal{V}$ represents patients thus remains the same across all graphs.
%All $K$ graphs share the set of vertices $V$.

\begin{algorithm}[t]
   \caption{Graph-based Similarity Feature Extraction}
   \label{graph}
\begin{algorithmic}
   \STATE {\bfseries Input:} Bipartite graphs $\{G_i\}_{i=1}^K$
   \STATE {\bfseries Output:} $\{\mathbf{X}_i\}_{i=1}^K$
   \FOR{$i=1$ {\bfseries to} $K$}
%   \STATE $G_c=\{U_c, V, E_c\}$, $U_c=\{u_i\}_{i=1}^{|U_c|}=\{u_i\}_{i=1}^{M_c}$
%   \STATE $E_c=\{\{w_{i,j}\}_{i=1}^{|U_c|}\}_{j=1}^{|V|}$, $V=\{v_j\}_{j=1}^{|V|}=\{v_j\}_{j=1}^{N}$
   \STATE $\mathcal{G}_i=\{\mathcal{U}_i, \mathcal{V}, \mathcal{E}_i\}, M=|\mathcal{U}_i|, N=|\mathcal{V}|$
   \STATE $\mathbf{A}=\begin{bmatrix} \mathbf{0}_{M,M} & \mathbf{B} \\ \mathbf{B}^{\text{T}} & \mathbf{0}_{N,N} \end{bmatrix}$ (where $\mathbf{B}[i,j]=w_{ij}$)
   \STATE $\mathbf{D}=\mathrm{diag}(\{\sum\limits_{j=1}^{M+N}\mathbf{A}[i,j]\}_{i=1}^{M+N})$
   \STATE $\mathbf{L}=\mathbf{D}^{-1/2}\mathbf{A}\mathbf{D}^{-1/2}$
   \STATE $\mathbf{X}_i=[\mathbf{e}_1,\mathbf{e}_2,\dots,\mathbf{e}_k]=\mathrm{eigendecomp}(\mathbf{L})$
   \FOR{$\alpha=1$ {\bfseries to} $M+N$}
   \STATE $\mathbf{X}_i[\alpha,:]=\mathbf{X}_i[\alpha,:] / (\sum_{\beta}\mathbf{X}_i[\alpha,\beta]^2)^{1/2}$
   \ENDFOR
   \ENDFOR
\end{algorithmic}
\end{algorithm}

Secondly, for each graph $\mathcal{G}_i$ we construct the adjacency matrix $\mathbf{A} \in \mathbb{R}^{(M+N)\times(M+N)}$.
%, of which each element is  $w_{i,cj}$.
%\cx{this "no interaction" assumption might not be that true. so we may need one sentence to justify it}
Since in $\mathcal{G}_i$ we ignore interaction within patients (or clinicians), $\mathbf{A}$ is block sparse, and takes the form as $\begin{bmatrix} \mathbf{0}_{M,M} & \mathbf{B} \\ \mathbf{B}^{\text{T}} & \mathbf{0}_{N,N} \end{bmatrix}$, where $\mathbf{B} \in \mathbb{R}^{M \times N}$ is the matrix representation of $\mathcal{E}_i$.

Thirdly, we compute the Laplacian matrix of $\mathcal{G}_i$ as $\mathbf{L}=\mathbf{D}^{-1/2}\mathbf{A}\mathbf{D}^{-1/2}$, where $\mathbf{D}$ is a diagonal matrix whose $(i,i)$-th element is the sum of the $i$-th row of $\mathbf{A}$.

Lastly, we compute the top $k$ eigenvectors of $\mathbf{L}$, i.e. $\mathbf{X}_i=[\mathbf{e}_1,\mathbf{e}_2,\dots,\mathbf{e}_k] \in \mathbb{R}^{(M+N) \times k}$, using algorithms such as \emph{Implicitly Restarted Arnoldi Method} \cite{lehoucq1996deflation} that is suitable to process large sparse matrices.
$\mathbf{X}_i$ is normalized such that each row of $\mathbf{X}_i$ has an $L_2$-norm of 1.

For a patient $p$, we extract the corresponding row from $\{\mathbf{X}_i\}_{i=1}^K$ as the relational similarity features. %$\{\mathbf{s}_i^p\}_{i=1}^K$.
% The $(M_c+i)$-th row of $\mathbf{X}$ can be used as the latent feature that measures the similarity of the $i$-th patient with respect to other patients in terms of visited clinicians, and will be combined with other types of features for predicting treatment initiation.
The complete algorithm is summarized in Algorithm \ref{graph}.

\section{Experiments and Results}
%We validate the effectiveness of the relational similarity feature by comparing the performance of several machine learning models with and without the proposed feature.

\subsection{Cohort}

%The task is to predict the initiation of first-line treatment for CLL patients using their clinical medical records.

We extract data from IQVIA longitudinal prescription (Rx) and medical claims (Dx) database, including hundreds of millions of patient clinical records.
In this study, we selected all the patients diagnosed with CLL from 01-2017 to 12-2018 from the IQVIA database and kept only the patients with complete Rx/Dx information.
We split the time period from 07-2017 to 12-2018 into 3 equal intervals.
Within each interval, we defined the positive cohort as patients who were diagnosed with CLL before the interval and started treatment within the interval, and the negative cohort as patients who were diagnosed with CLL before the interval, but did not start treatment during the interval.
The final positive and negative cohorts have 11,259 and 109,563 patient profiles, each of which contains a medical sequence and a feature vector including demographics and relational similarity features, respectively. 

In time series forecasting, it is common practice to reserve the last part of each time series for testing, and use the rest of the series for training \cite{bergmeir2012use}.
This is to avoid using information from the future to predict past events.
As a result, we cannot apply conventional train/test split of data such as k-fold cross-validation that would shuffle data randomly.
Therefore, the 07-2018 to 12-2018 interval is hold out for model testing, and the rest intervals are used for model training.

\subsection{Feature preparation}

\textbf{BOW features}: 
For each interval, we pulled one-year patient clinical records (diagnosis, procedure, etc.) before the interval. 
We divided the one-year look-back into four quarters, and used Bag-of-Words (BOW) \cite{mikolov2013efficient} to extract features (counts of occurrences of services) from the records within each quarter.
Medical services with less than 1\% occurrences in the entire dataset were ignored.
Eventually, we obtained 329 BOW features.

\textbf{Demographics and medical features}:
Demographic features include age and gender.
Medical features are 11 clinical services from patients' medical history, which are chosen by clinical experts and have been shown relevance to treatment initiation.
Both demographics and medical features are converted to categorical variables.

\textbf{Vectorized clinical time series}:
The medical service codes in the patients' clinical time series were converted to 300-dimensional dense vectors through a pre-trained embedding.

\textbf{Relational similarity}:
We constructed two bipartite graphs, one for patients and their diagnosis clinicians, and another for patients and their follow-up clinicians.
We took the top 5 eigenvectors from each of the two Laplacian matrices derived from the two graphs using Algorithm \ref{graph}, resulting in a 10-dimensional vector as the relational similarity feature for one patient.

\subsection{Evaluation}

% \cx{add evaluation strategy including data split, metrics , baselines, etc}
% \cx{add implementation details. Samples copied from my paper: For each method, the reported results (mean performance and its empirical standard deviation) are averaged over 20 independent runs. For each run, we randomly split the entire dataset into training (80\%), validation (10\%) and test sets (10\%). All models are built using the training and validation sets and then, evaluated using test set. We use Adam optimizer ~\cite{adam} to train each model, with the default learning rate set to 0.001. The number of training epoches for each model is set as 200 and an early stopping criterion is invoked if the performance does not improve in 20 epoches. All models are implemented in Keras with Tensorflow backend and tested on a system equipped with 64GB RAM, 12 Intel Core i7-6850K 3.60GHz CPUs and Nvidia GeForce GTX 1080.}

Because of the imbalance of positive and negative samples in the cohort, we use precision-recall area-under-curve (PR-AUC) and precision@k as the metrics to evaluate model performance.

% For each model, the reported results (mean performance and its empirical standard deviation) are averaged over 20 independent runs. 
% For each run, we randomly split the entire dataset into training (80\%), validation (10\%) and test sets (10\%). 

All models are built using the training data set and evaluated using the testing data set. 
We use Adam optimizer \cite{kingma2014adam} with a default learning rate of 1e-4. 
The number of training epochs for each model is 50 and an early stopping criterion is invoked if the performance does not improve in 10 consecutive epochs. 
All models are implemented in Keras with Tensorflow backend and tested on a system equipped with 128GB RAM, 16 Intel(R) Core Xeon(R) E5-2683 v4 2.10GHz CPUs, and Nvidia Tesla P100-PCIE-16GB.

\begin{figure}[t]
%\vskip 0.1in
\begin{center}
\centerline{\includegraphics[width=1.2\columnwidth]{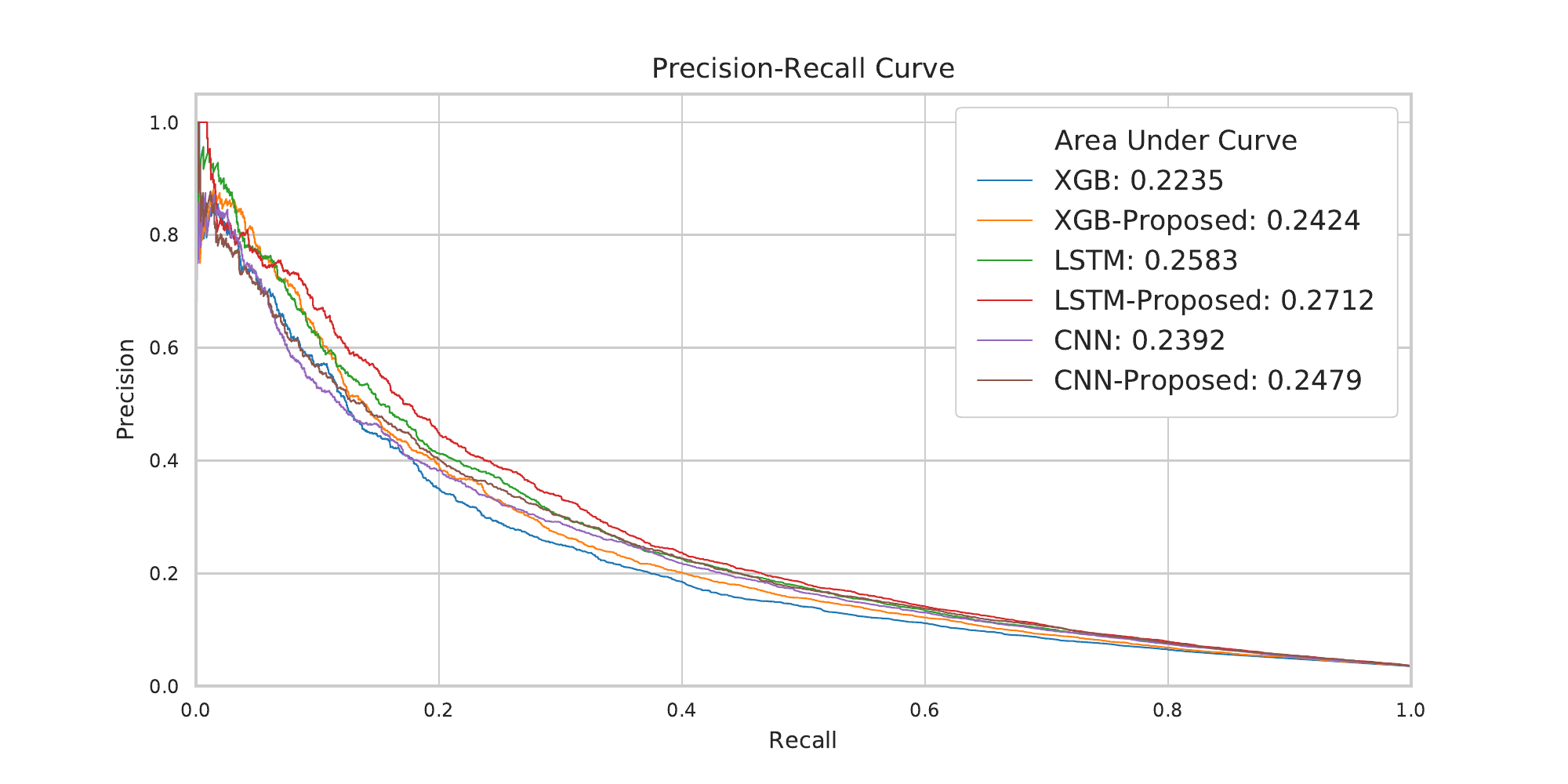}}
\caption{Comparison of baseline models with and without relational similarity feature measured in PR-AUC.}
\label{pr}
\end{center}
\vskip -0.4in
\end{figure}

\subsection{Models and Performance}\label{res}
To demonstrate the robustness of the proposed method, we have applied it on three baseline models for comparison: XGB, CNN, and LSTM. We evaluate the performance of these models with and without the relational similarity feature. For fairly comparison, we also introduce the diagnosis clinicians and follow-up clinicians IDs to the baseline models, to guarantee that the proposed method and baseline model leverage the same amount of information.

% \vspace{-7pt}
% \begin{itemize}
%     \item \textbf{XGB}: XGBoost model \cite{chen2016xgboost} with BOW features, demographic features, and medical features. No time-sensitive treatment was applied.
%     \item \textbf{CNN}: Same as XGB, except that a time-sensitive treatment was applied to the BOW features by dividing the one-year look-back evenly into 4 intervals (i.e. one quarter per interval).
%     \item \textbf{Doctor AI}: Same as XGB-TS, except that input features included similarity features as in equation \ref{feat2y}.
% \end{itemize} 
% \js{We should report clearly that numbers of features from each source}
\textbf{XGB}: 
A tree boosting regression model implemented in XGBoost with 500 estimators. Input features include BOW features, demographics, medical features, and relational similarity features.

\textbf{CNN}: 
A 4-layer CNN with a structure of $1\times1(128)-2\times2(128)-3\times3(128)-5\times5(128)$ (kernel size/number of kernels), followed by two dense layers.
Clinical time series are fed into the first CNN layer.
Features of relational similarity, demographics, and medical are concatenated with flattened CNN features at the first dense layer.

\textbf{LSTM}: 
A bi-directional LSTM handles the clinical time series.
The hidden dimension of the LSTM is 256.
The maximum hidden states at each time step of the top-level LSTM are concatenated with relational similarity features, demographics, and medical features, and processed by two dense layers.
%A customized Wide \& Deep structure \cite{cheng2016wide} with the deep component as an LSTM for handling clinical time series, and the wide component an a 2-layer feed-forward neural network) for handling relational similarity features, demographic features, and medical features.

\begin{figure}[t]
%\vskip 0.1in
\begin{center}
\centerline{\includegraphics[width=1.2\columnwidth]{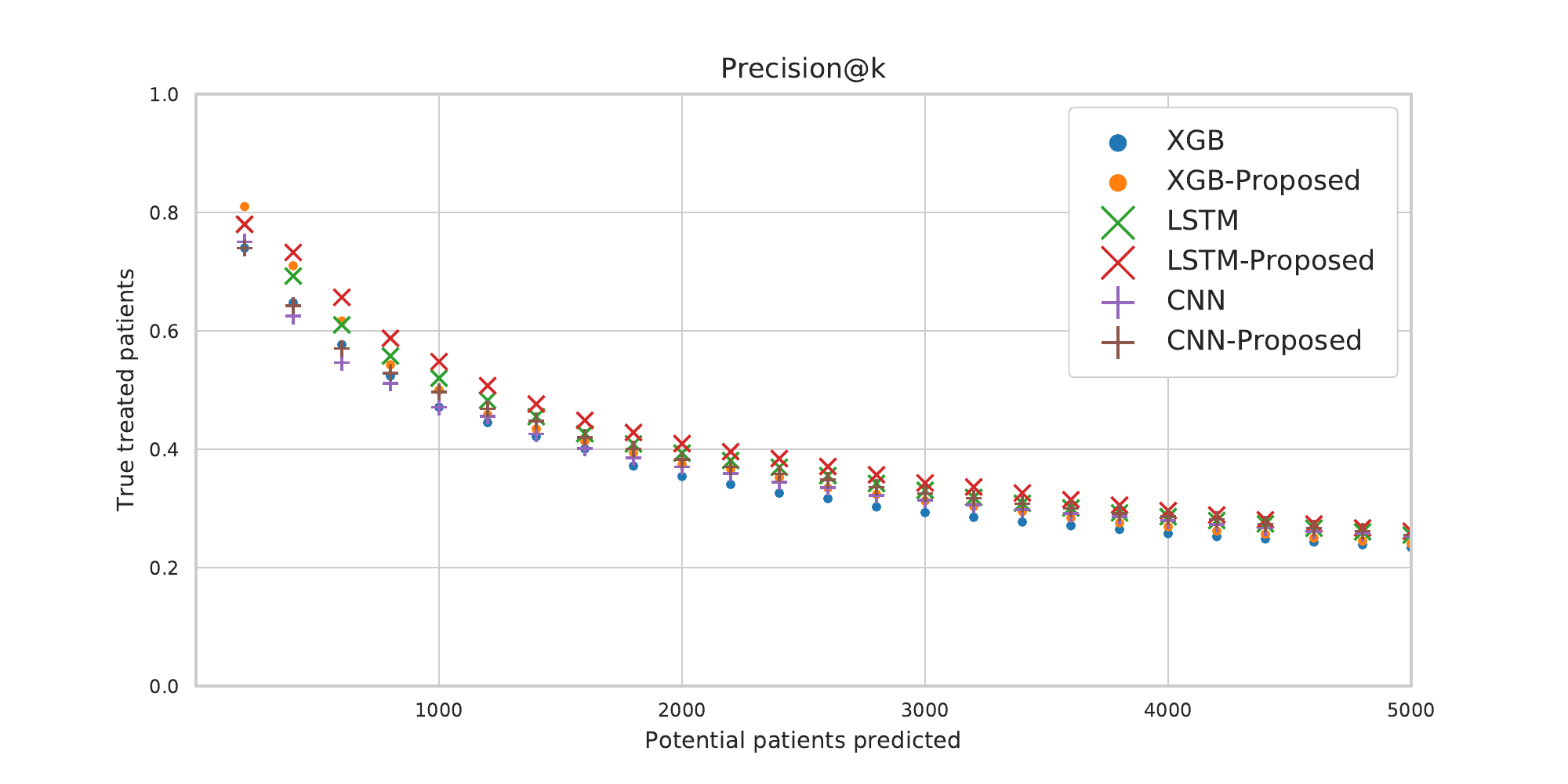}}
\caption{Comparison of baseline models with and without relational similarity feature measured in precision@k.}
\label{pk}
\end{center}
\vskip -0.2in
\end{figure}

Performance of the models with and without the relational similarity feature is evaluated in PR-AUC and precision@k, and shown in Figure \ref{pr} and \ref{pk}. Among all the models, the feature number of proposed and baseline are on comparable level, e.g., XGB-baseline has 1,331, and XGB-proposed has 1,339 features. 

% \js{so baselines have fewer features than ours? Then reviewers can argue of course the performance will be better as you used more features. It will be more convincing that we use the same number of features or fewer features to achieve better accuracy. e.g. what is performance if we only use the similarity features?}

A sampling of the precision@k curve is provided in Table \ref{pkkk}.
The results further confirm that by capturing the relational similarity feature, we can improve the prediction accuracy of all baseline models, suggesting the universal effectiveness of relational similarity feature for a wide range of model structures. When $K$ goes large, the proposed method consistently outperforms the baseline model, although the incremental gradually become marginal, e.g., in LSTM the improvement is 2.9\%, 1.6\%,  0.6\%, when $K$ equals 5,000, 10,000, 20,000, respectively. 

\section{Conclusions}
In this paper, we proposed a graph-based algorithm to extract latent relational similarity from patients' clinical time series.
%More notably, the algorithm for extracting similarity features is purely for feature engineering and can be used with arbitrary machine learning models beyond XGBoost.
Experimental results using real-world data show that the proposed feature can improve the prediction accuracy of a wide range of model structures.
We envision that the relational similarity can also enhance model performance on other tasks, such as rare disease detection or patient subtyping.
In its current form, the algorithm is operating on a two-dimensional feature space, i.e. patients and clinicians.
In the future, we will add medical services as another dimension into the feature space, and thus enable the application of more advanced signal processing techniques such as tensor decomposition to uncover more useful information from the multidimensional feature inputs.
% \js{The precision at K should be a probability score as oppose to the raw counts, and we need a sentence to justify the large K}

\begin{table}[t]
\centering
\caption{The number of successfully predicted treatments from 07-2018 to 12-2018 among the K selected patients.}
\vspace{3pt}
\label{pkkk}
\begin{tabular}{lrrrr}
\toprule[.8pt]
K & 600 & 1,200 & 1,800 & 3,600 \\ \midrule[.5pt]
XGB & 346 & 534 & 669 & 975 \\
XGB-Proposed & 370 & 550 & 709 & 1,023 \\
Improvement & 6.9\% & 3.0\% & 6.0\% & 4.9\% \\ 
\hline
CNN & 328 & 547 & 694 & 1,048 \\
CNN-Proposed & 342 & 562 & 721 & 1,080 \\
Improvement & 4.3\% & 2.7\% & 3.9\% & 3.1\% \\
\hline
LSTM & 366 & 579 & 736 & 1,083 \\
LSTM-Proposed & 394 & 609 & 771 & 1,133 \\
Improvement & 7.7\% & 5.2\% & 4.8\% & 4.6\% \\
\bottomrule[.8pt]
\end{tabular}
\vspace{-10pt}
\end{table}

% Hence for positive patients with $y=1$ who received the treatment, the patient's index date $t$ is set to be the day when the treatment was initiated; 
% For negative patients with $y=0$ who have not received the treatment, the index date $t$ is set to be the split date $t_{split}$, i.e. the most recent date in the database.

% Note use of \abovespace and \belowspace to get reasonable spacing
% above and below tabular lines.

% \begin{table}[t]
% \caption{Classification accuracies for naive Bayes and flexible
% Bayes on various data sets.}
% \label{sample-table}
% \vskip 0.15in
% \begin{center}
% \begin{small}
% \begin{sc}
% \begin{tabular}{lcccr}
% \toprule
% Data set & Naive & Flexible & Better? \\
% \midrule
% Breast    & 95.9$\pm$ 0.2& 96.7$\pm$ 0.2& $\surd$ \\
% Cleveland & 83.3$\pm$ 0.6& 80.0$\pm$ 0.6& $\times$\\
% Glass2    & 61.9$\pm$ 1.4& 83.8$\pm$ 0.7& $\surd$ \\
% Credit    & 74.8$\pm$ 0.5& 78.3$\pm$ 0.6&         \\
% Horse     & 73.3$\pm$ 0.9& 69.7$\pm$ 1.0& $\times$\\
% Meta      & 67.1$\pm$ 0.6& 76.5$\pm$ 0.5& $\surd$ \\
% Pima      & 75.1$\pm$ 0.6& 73.9$\pm$ 0.5&         \\
% Vehicle   & 44.9$\pm$ 0.6& 61.5$\pm$ 0.4& $\surd$ \\
% \bottomrule
% \end{tabular}
% \end{sc}
% \end{small}
% \end{center}
% \vskip -0.1in
% \end{table}

% Acknowledgements should only appear in the accepted version.
% \section*{Acknowledgements}

\end{document}